\documentclass[11pt,a4paper]{article}
\usepackage[hyperref]{acl2019}
\usepackage{times}
\usepackage{latexsym}

\usepackage{url}
\usepackage{graphicx}
\usepackage{amsthm}
\usepackage{amsfonts}
\usepackage{subcaption}
\usepackage{amsmath}
\usepackage{booktabs}
\usepackage{bbm}
\usepackage[linesnumbered,ruled]{algorithm2e}
\usepackage{mathtools}

\usepackage[noabbrev]{cleveref}
\usepackage{xcolor}
\usepackage{dirtytalk}

\newcommand{\argmax}{\mathop{\mathrm{argmax}}\limits}
\newcommand{\argmin}{\mathop{\mathrm{argmin}}\limits}
\newtheorem{theorem}{Theorem}

\aclfinalcopy % Uncomment this line for the final submission
\title{A Simple Theoretical Model of Importance for Summarization}

\author{Maxime Peyrard\thanks{Research partly done at UKP Lab from TU Darmstadt.}\\ %[.3em]
    EPFL\\
    \texttt{maxime.peyrard@epfl.ch}
}

\date{}

\begin{document}
\maketitle
\begin{abstract}
  Research on summarization has mainly been driven by empirical approaches, crafting systems to perform well on standard datasets with the notion of information \emph{Importance} remaining latent. We argue that establishing theoretical models of \emph{Importance} will advance our understanding of the task and help to further improve summarization systems. To this end, we propose simple but rigorous definitions of several concepts that were previously used only intuitively in summarization: \emph{Redundancy}, \emph{Relevance}, and \emph{Informativeness}. Importance arises as a single quantity naturally unifying these concepts. Additionally, we provide intuitions to interpret the proposed quantities and experiments to demonstrate the potential of the framework to inform and guide subsequent works.
\end{abstract}

\section{Introduction}
% \paragraph{Definition of Summarization}
Summarization is the process of identifying the most \emph{important information} from a source to produce a comprehensive output for a particular user and task \cite{Maybury:1999}.
While producing readable outputs is a problem shared with the field of \emph{Natural Language Generation}, the core challenge of summarization is the identification and selection of \emph{important information}. The task definition is rather intuitive but involves vague and undefined terms such as \emph{Importance} and \emph{Information}.

Since the seminal work of \newcite{Luhn}, automatic text summarization research has focused on empirical developments, crafting summarization systems to perform well on standard datasets leaving the formal definitions of \emph{Importance} latent \cite{das2007survey,Nenkova2012}.
This view entails collecting datasets, defining evaluation metrics and iteratively selecting the best-performing systems either via supervised learning or via repeated comparison of unsupervised systems \cite{Yao2017}.

Such solely empirical approaches may lack guidance as they are often not motivated by more general theoretical frameworks.
While these approaches have facilitated the development of practical solutions, they only identify signals correlating with the vague human intuition of \emph{Importance}. For instance, structural features like centrality and repetitions are still among the most used proxies for \emph{Importance} \cite{Yao2017,Kedzie}. However, such features just correlate with \emph{Importance} in standard datasets. Unsurprisingly, simple adversarial attacks reveal their weaknesses \cite{TUD-CS-2016-0154}.

% We postulate that establishing formal theories of \emph{Importance} will advance our understanding of the task and guide research in summarization. One can draw inspiration from physics, arguably one of the most successful scientific development, which fosters both empirical and theoretical works with strong interactions between the two. Empirical studies test hypothesis designed to falsify working theories, while theories are refined to account for new empirical results \cite{Kuhn:1970}. In summarization,
% the lack of efforts to produce abstract theoretical frameworks might impede the progress.

% A theory provides a frame of reference for interpreting observations, defining new concepts, generalizing knowledge and understanding complex logical relationships between variables. It forms an interrelated, coherent set of ideas and models which is refined upon new empirical observations \cite{Kuhn:1970}. Hence, it is, by design, more internally consistent than common sense and intuition.

% In symbiosis with empirical works, theories are particularly useful because they provide a common language to ground research. They describe how different approaches relate to each other, pinpoint dark zones and promising areas.
% Theoretically motivated experiments are always beneficial; even if the outcome of an experiment is unexpected, it is an opportunity to revise and improve the theory in a fundamental way \cite{Kuhn:1970}.

We postulate that theoretical models of \emph{Importance} are beneficial to organize research and guide future empirical works.
Hence, in this work, we propose a simple definition of information importance within an abstract theoretical framework.
This requires the notion of information, which has received a lot of attention since the work from \newcite{Shannon48} in the context of communication theory.
% The subsequent theory produced powerful tools applied successfully in various domains like
% physics \cite{PhysRev.106.620},
% economics \cite{EcoIT},
% evolutionary biology \cite{NYAS:NYAS6422},
% or even the study of consciousness \cite{IIT}.
Information theory provides the means to rigorously discuss the abstract concept of information, which seems particularly well suited as an entry point for a theory of summarization.
However, information theory concentrates on uncertainty (entropy) about which message was chosen from a set of possible messages, ignoring the semantics of messages \cite{Shannon48}.
Yet, summarization is a lossy semantic compression depending on background knowledge.

% However, information theory focused on uncertainty (entropy) about which message was chosen among a set of possible messages, ignoring the semantic of messages \cite{Shannon48}. Shannon's theory offers tools for lossless syntactic compression of messages disregarding background knowledge, but summarization is a lossy semantic compression depending on background knowledge.

In order to apply information theory to summarization, we assume texts are represented by probability distributions over so-called \emph{semantic units} \cite{bao2011towards}. This view is compatible with the common distributional embedding representation of texts rendering the presented framework applicable in practice. When applied to semantic symbols, the tools of information theory indirectly operate at the semantic level \cite{Carnap,Zhong}. \\

\noindent
\textbf{Contributions}: \\
\begin{itemize}
\item We define several concepts intuitively connected to summarization: \emph{Redundancy}, \emph{Relevance} and \emph{Informativeness}. These concepts have been used extensively in previous summarization works and we discuss along the way how our framework generalizes them.

\item From these definitions, we formulate properties required from a useful notion of \emph{Importance} as the quantity unifying these concepts. We provide intuitions to interpret the proposed quantities.

\item Experiments show that, even under simplifying assumptions, these quantities correlates well with human judgments making the framework promising in order to guide future empirical works.
\end{itemize}
% In fact, the \emph{Importance} of a semantic unit depends on which other units are present within some contextual boundaries: \emph{Redundancy} in the context of the summary only, \emph{Relevance} in the context of the source document(s) and \emph{Informativeness} in the context of background knowledge and preconceptions of the user.

% \emph{Importance} encompasses these three levels.
% Finally, whenever one compresses with loss of information one must make choices about what to discard. \emph{Importance} is the measure that guides these choices.

% We also provide insights and examples to interpret the proposed quantities.
% Apart from providing a theoretical model of summarization, our work opens-up many promising research directions which can directly improve summarization systems and evaluation metrics.

% \section{Background}
% \label{sec:background}
% \input{background}

\section{Framework}
\label{sec:approach}
\subsection{Terminology and Assumptions}
We call \emph{semantic unit} an atomic piece of information
\cite{Zhong,cruse86lexicalsemantics}. We note \(\Omega\) the set of all possible semantic units.

A text $X$ is considered as a semantic source emitting semantic units as envisioned by \newcite{weaver1953recent} and  discussed by \newcite{bao2011towards}. Hence, we assume that $X$ can be represented by a probability distribution $\mathbb{P}_X$ over the semantic units $\Omega$. \\

\noindent
\textbf{Possible interpretations}: \\
One can interpret $\mathbb{P}_X$ as the frequency distribution of semantic units in the text. Alternatively, $\mathbb{P}_X(\omega_i)$ can be seen as the (normalized) likelihood that a text $X$ entails an atomic information $\omega_i$ \cite{Carnap}. Another interpretation is to view $\mathbb{P}_X(\omega_i)$ as the normalized contribution (utility) of $\omega_i$ to the overall meaning of $X$ \cite{Zhong}. \\

% For practical considerations, these interpretations are equivalent, but we discuss semantic units in more details in the Discussion section.

\noindent
\textbf{Motivation for semantic units}: \\
In general, existing semantic information theories either postulate or imply the existence of semantic units \cite{Carnap,bao2011towards,Zhong}. For example, the \emph{Theory of Strongly Semantic Information} produced by \newcite{floridi2009philosophical} implies the existence of semantic units (called information units in his work). Building on this, \newcite{tsvetkov2014ke} argued that the original theory of Shannon can operate at the semantic level by relying on semantic units.

In particular, existing semantic information theories imply the existence of semantic units in formal semantics \cite{Carnap}, which treat natural languages as formal languages \cite{Montague1970-MONEAA-2}.
In general, lexical semantics \cite{cruse86lexicalsemantics} also postulates the existence of elementary constituents called minimal semantic constituents. For instance, with frame semantics \cite{Fillmore}, frames can act as semantic units.

Recently, distributional semantics approaches have received a lot of attention \cite{Turian:2010,word2vec}. They are based on the distributional hypothesis \cite{harris54} and the assumption that meaning can be encoded in a vector space \cite{turney2010frequency,erk2010word}. These approaches also search latent and independent components that underlie the behavior of words \cite{gabor-EtAl,abs-1301-3781}.

While different approaches to semantics postulate different basic units and different properties for them, they have in common that \emph{meaning arises from a set of independent and discrete units}. Thus, the semantic units assumption is general and has minimal commitment to the actual nature of semantics. This makes the framework compatible with most existing semantic representation approaches. Each approach specifies these units and can be plugged in the framework, e.g., frame semantics would define units as frames, topic models \cite{Allahyari:2017} would define units as topics and distributional representations would define units as dimensions of a vector space. \\

In the following paragraphs, we represent the source document(s) \(D\) and a candidate summary \(S\) by their respective distributions \(\mathbb{P}_{D}\) and \(\mathbb{P}_{S}\).\footnote{We sometimes note $X$ instead of $\mathbb{P}_{X}$ when it is not ambiguous}
% Then, we propose intuitive definitions for \emph{Redundancy}, \emph{Relevance} and \emph{Informativeness}, by relying on the well-established field of information theory to provide sound theoretical motivations.
% \emph{Importance} arises a the quantity that satisfies simple requirements formulated from the . We notice that it can be interpreted in terms of the other intuitive quantities.
% Since we introduced semantic symbols, the information theoretic tools indirectly operate at the semantic level.

\subsection{Redundancy}
\label{ssec:redundancy}
Intuitively, a summary should contain a lot of information.
In information-theoretic terms, the \emph{amount of information} is measured by Shannon's entropy. For a summary \(S\) represented by \(\mathbb{P}_{S}\):
\begin{equation}
H(S) = - \sum\limits_{\omega_i} \mathbb{P}_{S}(\omega_i) \cdot \log(\mathbb{P}_{S}(\omega_i))
\end{equation}
% where \(\mathbb{P}_{S}(\omega_j)\) is simply the frequency of the semantic unit \(\omega_i\) in the text \(S\).

\(H(S)\) is maximized for a uniform probability distribution when every semantic unit is present only once in \(S\): \(\forall (i,j),  \mathbb{P}_{S}(\omega_i) = \mathbb{P}_{S}(\omega_j)\). Therefore, we define \emph{Redundancy}, our first quantity relevant to summarization, via entropy:
\begin{equation}
Red(S) = H_{max} - H(S)
\end{equation}
Since $H_{max} = \log |\Omega|$ is a constant indepedent of $S$, we can simply write: $Red(S) = -H(S)$. \\

% Intuitively, a summary \(S\) maximizes the information content if it displays many semantic units but once.
% Indeed, \(S\) should not be redundant but also contain as many semantic units as possible. The two following summaries: \(S_1 = (a,b)\) and \(S_2 = (a,b,c)\) are both non-redundant but $S_2$ is intuitively better because it contains more information. This is captured by entropy because \(H(S_1) =  \log(2) \le H(S_2) = \log(3)\).
% Furthermore,
\noindent
\textbf{Redundancy in Previous Works}:\\
By definition, entropy encompasses the notion of maximum coverage. Low redundancy via maximum coverage is the main idea behind the use of submodularity \cite{LinB11}. Submodular functions are generalizations of coverage functions which can be optimized greedily with guarantees that the result would not be far from optimal \cite{opac-b1135069}. Thus, they have been used extensively in summarization \cite{Sipos2012, yogatama-liu-smith:2015:EMNLP}. Otherwise, low redundancy is usually enforced during the extraction/generation procedures like MMR \cite{Carbonell:1998}.

% Indeed, maximum coverage and minimum redundancy are equivalxent in the context of summarization because one should fit as much information as possible in a constrained space \cite{McDonald2007}.

% Now, suppose \(T\) displays the following semantic units: \((a,b,a,c,b)\). Then the summaries: \((a,b)\), \((b,c)\) and \((a,c)\) are all maximizing \(H\), minimizing \emph{Redundancy} and thus indistinguishable without further insights.

\subsection{Relevance}
\label{ssec:relevance}
% A summary also depends on the source(s) it originates from.
Intuitively, observing a summary should reduce our uncertainty about the original text. A summary approximates the original source(s) and this approximation should incur a minimum loss of information. This property is usually called \emph{Relevance}.

% With information theoretic tools, estimating \emph{Relevance} boils to compare the distribution $S$ and $T$. The surprise of seeing $\omega_i$ is given by \(-\log(\mathbb{P}_T(\omega_i))\) and therefore the average surprise of observing $S$ while knowing $T$ is given by the cross-entropy $Rel(S,T) = CE(S,T)$:

Here, estimating \emph{Relevance} boils down to comparing the distributions $\mathbb{P}_S$ and $\mathbb{P}_D$, which is done via the cross-entropy $Rel(S,D) = -CE(S,D)$:

\begin{equation}
Rel(S,D) = \sum\limits_{\omega_i} \mathbb{P}_S(\omega_i) \cdot \log(\mathbb{P}_D(\omega_i))
\end{equation}
The cross-entropy is interpreted as the average surprise of observing $S$ while expecting $D$. A summary with a low expected surprise produces a low uncertainty about what were the original sources. This is achieved by exhibiting a distribution of semantic units similar to the one of the source documents: \(\mathbb{P}_S \approx \mathbb{P}_D\).
% The cross-entropy also measure the amount of new information introduced by the summary $S$ given that we already know $T$. In order to reduce the uncertainty about $T$ a summary $S$ should introduce no new information and strick $T$.

Furthermore, we observe the following connection with \emph{Redundancy}:
\begin{equation}
\begin{split}
KL(S||D) = CE(S,D) - H(S) \\
-KL(S||D) = Rel(S,D) - Red(S)
\end{split}
\end{equation}
KL divergence is the information loss incurred by using $D$ as an approximation of $S$ (i.e., the uncertainty about $D$ arising from observing $S$ instead of $D$). A summarizer that minimizes the KL divergence minimizes \emph{Redundancy} while maximizing \emph{Relevance}.

In fact, this is an instance of the \emph{Kullback Minimum Description Principle} (MDI) \cite{kullback1951}, a generalization of the \emph{Maximum Entropy Principle} \cite{PhysRev.106.620}: the summary minimizing the KL divergence is the least biased (i.e., least redundant or with highest entropy) summary matching $D$. In other words, this summary fits $D$ while inducing a minimum amount of \emph{new} information. Indeed, any \emph{new} information is necessarily biased since it does not arise from observations in the sources. The MDI principle and KL divergence unify \emph{Redundancy} and \emph{Relevance}. \\

\noindent
\textbf{Relevance in Previous Works}:\\
\emph{Relevance} is the most heavily studied aspect of summarization. In fact, by design, most unsupervised systems model \emph{Relevance}. Usually, they used the idea of \emph{topical frequency} where the most frequent topics from the sources must be extracted. Then, different notions of \emph{topics} and counting heuristics have been proposed. We briefly discuss these developments here.

\newcite{Luhn} introduced the simple but influential idea that sentences containing the most important words are most likely to embody the original document. Later, \newcite{Nenkova:2006:CCS} showed experimentally that humans tend to use words appearing frequently in the sources to produce their summaries. Then, \newcite{VANDERWENDE20071606} developed the system \emph{SumBasic}, which scores each sentence by the average probability of its words.

The same ideas can be generalized to n-grams. A prominent example is the ICSI system \cite{Gillick2009} which extracts frequent bigrams. Despite being rather simple, ICSI produces strong and still close to state-of-the-art summaries \cite{HONG14}.

Different but similar words may refer to the same topic and should not be counted separately. This observation gave rise to a set of important techniques based on topic models \cite{Allahyari:2017}. These approaches cover sentence clustering \cite{McKeown:1999:TMS,Radev2000,zhang:2015:NAACL}, lexical chains \cite{barzilay1999using}, Latent Semantic Analysis \cite{deerwester1990indexing} or Latent Dirichlet Allocation \cite{blei2003latent} adapted to summarization \cite{hachey2006dimensionality, daume2006bayesian, wang2009multi, davis2012occams}.
Approaches like hLDA can exploit repetitions both at the word and at the sentence level \cite{celikyilmaz:2010:ACL}.

Graph-based methods form another particularly powerful class of techniques to estimate the frequency of topics, e.g., via the notion of centrality \cite{Mani1997,mihalcea2004textrank,Erkan2004}.
A significant body of research was dedicated to tweak and improve various components of graph-based approaches. For example, one can investigate different similarity measures \cite{chali2008improving}. Also, different weighting schemes between sentences have been investigated \cite{leskovec2005impact,wan2006improved}.

Therefore, in existing approaches, the topics (i.e., atomic units) were words, n-grams, sentences or combinations of these. The general idea of preferring \emph{frequent topics} based on various counting heuristics is formalized by cross-entropy. Indeed, requiring the summary to minimize the cross-entropy with the source documents implies that frequent topics in the sources should be extracted first.

An interesting line of work is based on the assumption that the best sentences are the ones that permit the best reconstruction of the input documents \cite{he2012document}. It was refined by a stream of works using distributional similarities \cite{Li:2015:RMS,Liu:2015:MSB,ma:2016:COLING}. There, the atomic units are the dimensions of the vector spaces. This information bottleneck idea is also neatly captured by the notion of cross-entropy which is a measure of information loss. Alternatively, \citep{Daume:2002} viewed summarization as a noisy communication channel which is also rooted in information theory ideas. \citep{Relevance-theory} provide a more general and less formal discussion of relevance in the context of Relevance Theory \cite{lavrenko2008generative}.

\subsection{Informativeness}
\label{ssec:informativeness}
% In Shannon's theory, information is the surprise of observing an outcome. The surprise depends on our expectations and preconceptions about the outcome, which is encoded in a probability distribution $P$ over symbols.
% When extending information to the notion of semantic information, the reasoning is analogous. Semantic information is the surprise of observing a semantic signal \cite{Carnap, Zhong}. This surprise depends on our expectation and preconceptions about which semantic signal will happen. Therefore we introduce $K$, which represents background knowledge and preconceptions about the summarization task. In the classical view \(P\) is a probability distribution over symbols, and analogously \(K\) has a probability distribution \(\mathbb{P}_K\) over semantic units \(\Omega\).

\emph{Relevance} still ignores other potential sources of information such as previous knowledge or preconceptions. We need to further extend the contextual boundary.
Intuitively, a summary is informative if it induces, for a user, a great change in her knowledge about the world.
% Intuitively, a summary is informative if it gives a lot of new information with respect to what is already known. An informative summary is a summary that induces, for a user, a great change in his/her knowledge about the world.
Therefore, we introduce $K$, the background knowledge (or preconceptions about the task). $K$ is represented by a probability distribution $\mathbb{P}_K$ over semantic units $\Omega$.

Formally, the amount of \emph{new} information contained in a summary $S$ is
% the average amount of surprise with respect to what we know. The surprise of seeing $\omega_i$ is given by \(-\log(\mathbb{P}_K(\omega_i))\) and therefore the average surprise of observing $S$ while knowing $K$ is
given by the cross-entropy $Inf(S,K) = CE(S,K)$:
\begin{equation}
Inf(S, K) = - \sum\limits_{\omega_i} \mathbb{P}_S(\omega_i) \cdot \log(\mathbb{P}_K(\omega_i))
\end{equation}
For \emph{Relevance} the cross-entropy between $S$ and $D$ should be low. However, for \emph{Informativeness}, the cross-entropy between $S$ and $K$ should be high because we measure the amount of new information induced by the summary in our knowledge.
% In the next section, we will unify both principles, which means that a summary should, by using only information available in $T$ produce what brings the most new information to a user with previous background $K$.

% \emph{Informativeness} is also connected to entropy via KL divergence:
% \begin{equation}
% \begin{split}
% KL(S||K) = CE(S,K) - H(S) \\
% KL(S||K) = Inf(S, K) - Red(S)
% \end{split}
% \end{equation}
% KL maximization unifies \emph{Redundancy} and \emph{Informativeness}. \\

% To clarify, $Inf(S,K)$ states that a summary should induce new information with respect to $K$, which would be the most biased summary according to the Kullback MDI principle. In the next section, by unifying $Inf$ and $Rel$, that a summary should include new information with respect to $K$ but use only information coming from $T$

% \paragraph{Remark}
Background knowledge is modeled by assigning a high probability to known semantic units. These probabilities correspond to the strength of $\omega_i$ in the user's memory. A simple model could be the uniform distribution over known information: \(\mathbb{P}_K(\omega_i)\) is \(\frac{1}{n}\) if the user knows \(\omega_i\), and $0$ otherwise.
        % However, $K$ is more general than just background knowledge. It can control many variants of summarization tasks.
        However, $K$ can control other variants of the summarization task:
        A personalized $K_p$ models the preferences of a user by setting low probabilities to the semantic units of interest.
        Similarly, a query $Q$ can be encoded by setting low probability to semantic units related to $Q$.
        Finally, there is a natural formulation of update summarization. Let \(U\) and \(D\) be two sets of documents. Update summarization consists in summarizing \(D\) given that the user has already seen \(U\). This is modeled by setting \(K = U\), considering \(U\) as previous knowledge. \\

% One could also define \emph{Informativeness} with respect to several \(K\)'s. For example, it is possible to define background knowledge \(K\), a query \(Q\) and background documents \(U\). An overall \emph{Informativeness} would be derived from an aggregated $K$.

\noindent
\textbf{Informativeness in Previous Works}:\\
The modelization of \emph{Informativeness} has received less attention by the summarization community.
The problem of identifying stopwords originally faced by \newcite{Luhn} could be addressed by developments in the field of information retrieval using background corpora like TF$\cdot$IDF \cite{sparck1972statistical}. Based on the same intuition, \newcite{dunning1993accurate} outlined an alternative way of identifying highly descriptive words: the \emph{log-likelihood ratio} test. Words identified with such techniques are known to be useful in news summarization \cite{Harabagiu:2005}.

Furthermore, \newcite{conroy-schlesinger-oleary:2006} proposed to model background knowledge by a large random set of news articles. In update summarization, \newcite{Delort:2012} used Bayesian topic models to ensure the extraction of informative summaries. \newcite{louis:2014:P14-2} investigated background knowledge for update summarization with Bayesian surprise. This is comparable to the combination of \emph{Informativeness} and \emph{Redundancy} in our framework when semantic units are n-grams. Thus, previous approaches to \emph{Informativeness} generally craft an alternate background distribution to model the \emph{a-priori} importance of units. Then, units from the document rare in the background are preferred, which is captured by maximizing the cross-entropy between the summary and $K$. Indeed, unfrequent units in the background would be preferred in the summary because they would be surprising (i.e., informative) to an average user.

\subsection{Importance}
\label{ssec:imp}
Since \emph{Importance} is a measure that guides which choices to make when discarding semantic units, we must devise a way to encode their relative importance.
Here, this means finding a probability distribution unifying $D$ and $K$ by encoding expectations about which semantic units should appear in a summary.

    \emph{Informativeness} requires a biased summary (w.r.t. $K$) and \emph{Relevance} requires an unbiased summary (w.r.t. $D$). Thus, a summary should, by using only information available in $D$, produce what brings the most new information to a user with knowledge $K$. This could formalize a common intuition in summarization that units frequent in the source(s) but rare in the background are important.
    % In other words, the bias w.r.t. $K$ should be informed by the sources $D$.

    Formally, let $d_i = \mathbb{P}_D(\omega_i)$ be the probability of the unit $\omega_i$ in the source $D$. Similarly, we note $k_i = \mathbb{P}_K(\omega_i)$. We seek a function $f(d_i, k_i)$ encoding the importance of unit $\omega_i$.
    We formulate simple requirements that $f$ should satisfy:
    % From the previous insights,
    % we formulate simple requirements that the function $f$ should satisfy:
    \begin{itemize}
    \item Informativeness: $\forall i \neq j$, if $d_i = d_j$ and $k_i > k_j$ then $f(d_i, k_i) < f(d_j, k_j)$
    \item Relevance: $\forall i \neq j$, if $d_i > d_j$ and $k_i = k_j$ then $f(d_i, k_i) > f(d_j, k_j)$
    \item Additivity: $I(f(d_i, k_i)) \equiv \alpha I(d_i) + \beta I(k_i)$ ($I$ is the information measure from Shannon's theory \cite{Shannon48})
    \item Normalization: $\sum\limits_{i} f(d_i, k_i) = 1$
    \end{itemize}
    The first requirement states that, for two semantic units equally represented in the sources, we prefer the more informative one. The second requirement is an analogous statement for \emph{Relevance}. The third requirement is a consistency constraint to preserve additivity of the information measures \cite{Shannon48}.
    The fourth requirement ensures that $f$ is a valid distribution.

    \begin{theorem}
    \label{th:main_th}
      The functions satisfying the previous requirements are of the form:
      \begin{align}
          \mathbb{P}_{\frac{D}{K}}(\omega_i) & = \frac{1}{C} \cdot \frac{d_i^{\alpha}}{k_i^{\beta}} \\
          C & = \sum\limits_{i} \frac{d_i^{\alpha}}{k_i^{\beta}} \text{ , } \ \alpha, \beta \in \mathbb{R}^{+}
      \end{align}
      % For simplicity, we now use $(\alpha, \beta) = (1,1)$.
      \end{theorem}
      $C$ is the normalizing constant. The parameters $\alpha$ and $\beta$ represent the strength given to \emph{Relevance} and \emph{Informativeness} respectively which is made clearer by \cref{eq:decomp_ik}. The proof is provided in \cref{sec:proof}. \\

% If $\beta \to 0$, then $\mathbb{P}_{\frac{T}{K}} \to \mathbb{P}_{T}$ and we recover \emph{Relevance}. Similarly, when $\alpha \to 0$, we recover \emph{Informativeness}.
% In the next paragraph, we clarify the role of $\mathbb{P}_{\frac{T}{S}}$ as a target distribution.

% The distribution $\mathbb{P}_{\frac{T}{K}}$ is central because it encodes the relative importance of semantic units and gives an overall target for the summary. $I_K$ states that a summary should approximate this distribution.

    \noindent
    \textbf{Summary scoring function}: \\
        By construction, a candidate summary should approximate $\mathbb{P}_{\frac{D}{K}}$, which encodes the relative importance of semantic units. Furthermore, the summary should be non-redundant (i.e., high entropy).
        These two requirements are unified by the Kullback MDI principle: The least biased summary $S^{*}$ that best approximates the distribution $\mathbb{P}_{\frac{D}{K}}$ is the solution of:
        \begin{equation}
        S^{*} = \argmax\limits_{S} \theta_I = \argmin\limits_{S} KL(S || \mathbb{P}_{\frac{D}{K}})
        \end{equation}
        % The Kullback MDI principle selects the summary $S^{*}$ which fits the expectation while making no further commitment (minimum redundancy, low bias).
        % Therefore, we say that $\theta_I$ measures the quality of a summary. \\
    % This is further clarified in the following paragraph.
        Thus, we note $\theta_I$ as the quantity that scores summaries:
        \begin{equation}
            \theta_I(S, D, K) = - KL(\mathbb{P}_S, || \mathbb{P}_{\frac{D}{K}})
        \end{equation}

        % \textbf{Remark}: We note that a summary maximizing the \emph{Relevance} also follows the Kullback MDI and is the least biased summary fiting the expectations encoded by $D$. Similary, a summary maximizing the \emph{Informativeness} is an instance of the Kullback MDI where the expectations are given by $\frac{1}{K}$. Trivially, a summary minimizing its redundancy (maximizing its entropy) is an instance of the maximum entropy principle. Thus, maximum entropy is the driving force behind the formalization.\\

    \noindent
    \textbf{Interpretation of $\mathbb{P}_{\frac{D}{K}}$}:\\
    % In this chapter, we clarify the role of the distribution $\mathbb{P}_{\frac{T}{K}}$ and provide intuitions to interpret it.
        $\mathbb{P}_{\frac{D}{K}}$ can be viewed as an \emph{importance-encoding distribution} because it encodes the relative importance of semantic units and gives an overall target for the summary.
        % The Kullback MDI principle states that a summary should approximate this distribution by minimizing the KL divergence.

        For example, if a semantic unit $\omega_i$ is prominent in $D$ ($\mathbb{P}_D(\omega_i)$ is high) and not known in $K$ ($\mathbb{P}_D(\omega_i)$ is low), then $\mathbb{P}_{\frac{D}{K}}(\omega_i)$ is very high, which means very desired in the summary. Indeed, choosing this unit will fill the gap in the knowledge $K$ while matching the sources.

\begin{figure*}
        \centering
        \begin{subfigure}{.33\textwidth}
          \centering
          \includegraphics[width=\linewidth]{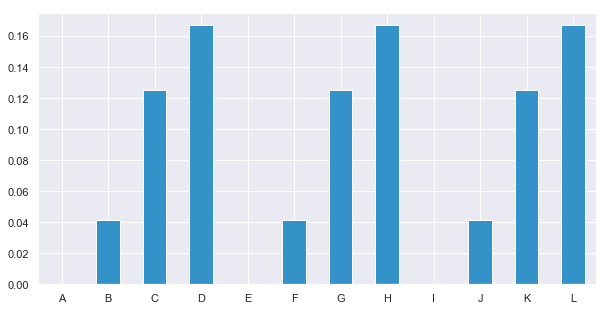}
          \caption{ditribution $\mathbb{P}_{D}$}
          \label{fig:sub1}
        \end{subfigure}%
        \begin{subfigure}{.33\textwidth}
          \centering
          \includegraphics[width=\linewidth]{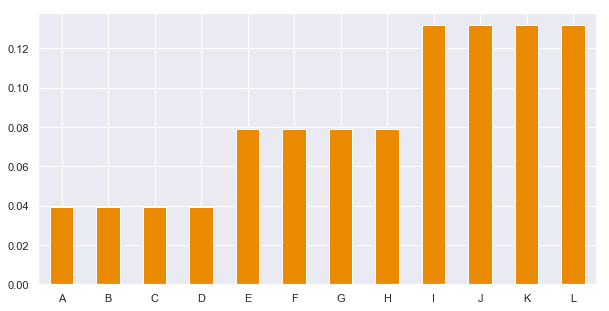}
          \caption{distribution $\mathbb{P}_{K}$ }
          \label{fig:sub2}
        \end{subfigure}
        \begin{subfigure}{.33\textwidth}
          \centering
          \includegraphics[width=\linewidth]{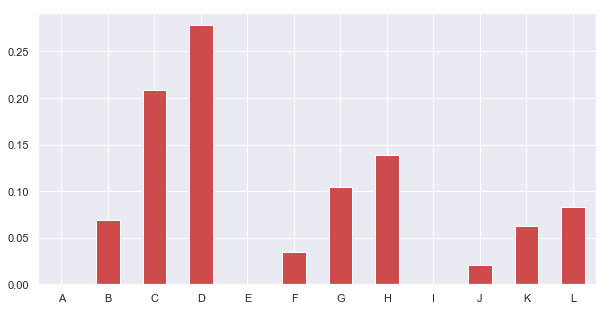}
          \caption{distribution $\mathbb{P}_{\frac{D}{K}}$}
          \label{fig:sub3}
        \end{subfigure}
        \caption{\cref{fig:sub1} represents an example distribution of sources, \cref{fig:sub2} an example distribution of background knowledge and \cref{fig:sub3} is the resulting target distribution that summaries should approximate.}
        \label{fig:visu_t_k}
        \end{figure*}

        \Cref{fig:visu_t_k} illustrates how this distribution behaves with respect to $D$ and $K$ (for $\alpha = \beta = 1$).
        % Depending on their prominence in the sources and background knowledge, $\mathbb{P}_{\frac{T}{K}}(\omega_i)$ gives the relative importance of semantic units.
        \\

    \noindent
    \textbf{Summarizability}: \\
        The target distribution $\mathbb{P}_{\frac{D}{K}}$ may exhibit different properties. For example, it might be clear which semantic units should be extracted (i.e., a spiky probability distribution) or it might be unclear (i.e., many units have more or less the same importance score). This can be quantified by the entropy of the importance-encoding distribution:
        \begin{equation}
          H_{\frac{D}{K}} = H(\mathbb{P}_{\frac{D}{K}})
        \end{equation}
        Intuitively, this measures the number of possibly good summaries. If $H_{\frac{D}{K}}$ is low then $\mathbb{P}_{\frac{D}{S}}$ is spiky and there is little uncertainty about which semantic units to extract (few possible \emph{good} summaries). Conversely, if the entropy is high, many equivalently \emph{good} summaries are possible. \\

    \noindent
    \textbf{Interpretation of $\theta_I$}: \\
        % Maximizing $\theta_I$ not only encourages the selection of high scoring semantic units, it also indicates which choices and trade-offs are more beneficial.
        % $\theta_I$ covers the overall selection of several semantic units together, it is not restricted to independently choosing individual ones.
        To better understand $\theta_I$, we remark that it can be expressed in terms of the previously defined quantities:
        \begin{align}
        \label{eq:decomp_ik}
            % \theta_I(S, D, K) & = H(S) - \alpha CE(S, D)\\
            %                   & \ \ + \beta CE(S, K) + \log C\\
            \theta_I(S, D, K) & \equiv -Red(S) + \alpha Rel(S,D) \\
                                & \ \  + \beta Inf(S,K)
        \end{align}
        Equality holds up to a constant term $\log C$ independent from $S$.
        % This gives t interpretations of $I_K$:
        % \begin{align}
        %     I_K(S;D) & \equiv -Red(S) + \alpha Rel(S,D) + \beta Inf(S,K) \\
        %     I_K(S;D) & \equiv KL(S,D) + Inf(S, K) \\
        %     I_K(S;D) & \equiv Rel(S,D) + KL(S, K)
        % \end{align}
        Maximizing $\theta_I$ is equivalent to maximizing \emph{Relevance} and \emph{Informativeness} while minimizing \emph{Redundancy}. Their relative strength are encoded by $\alpha$ and $\beta$.

        % The second line says that maximizing $I_K$ is equivalent to maximizing both \emph{Relevance} and \emph{Informativeness}.

        % The third says that it is equivalent to minimizing the average surprise of observing $S$ expecting $D$ while maximizing the Bayesian surprise of observing $S$ knowing $K$. The third one is the minimization of \emph{Redundancy} while maximizing both \emph{Relevance} and Bayesian surprise.
        Finally, $H(S)$, $CE(S,D)$ and $CE(S,K)$ are the three independent components of \emph{Importance}. \\

        It is worth noting that each previously defined quantity: $Red$, $Rel$ and $Inf$ are measured in bits (using base $2$ for the logarithm). Then, $\theta_I$ is also an information measure expressed in bits. \newcite{Shannon48} initially axiomatized that information quantities should be additive and therefore $\theta_I$ arising as the sum of other information quantities is unsurprising. Moreover, we ensured additivity with the third requirement of $\mathbb{P}_{\frac{D}{K}}$.

%% #############
%% ############# Universality

% \paragraph{Generality of $I_K$}
% Let $\sigma$ be a summarizer. $\sigma$ is a set function which maps a text $T$ to another shorter text $S$: $\sigma(T) = S$.
% With this broad definition, many summarizers are possible. Indeed, if $n$ is the number of source texts and $m$ the number of summaries available, then there are $m^n$ summarizers. Most of them do not exhibit pattern or structure, meaning that two very similar input texts $T_1 \approx T_2$ could be mapped to completely different summaries $\sigma(T_1)$ and $\sigma(T_2)$.

% We may ask ourselve, which summarizers can be modeled by $I_K$, or for which summarizer $\sigma$ there exists $\alpha_{\sigma}, \beta_{\sigma}$ and $K_{\sigma}$, such that:
% \begin{equation}
% \begin{split}
% \forall T, \sigma(T) = \argmin\limits_{S} I_K \\
% 					= \argmin\limits_{S} H(S) - \alpha CE(S,T) + \beta CE(S,K)
% \end{split}
% \end{equation}

%% #############
%% ############# Summarization Uncertainty

\subsection{Potential Information}
\emph{Relevance} relates $S$ and $D$, \emph{Informativeness} relates $S$ and $K$, but we can also connect $D$ and $K$.
Intuitively, we can extract a lot of new information from $D$ only when $K$ and $D$ are different.

With the same argument laid out for \emph{Informativeness}, we can define the amount of potential information as the average surprise of observing $D$ while already knowing $K$. Again, this is given by the cross-entropy $PI_K(D) = CE(D, K)$:
\begin{equation}
    PI_K(D) = - \sum\limits_{\omega_i} \mathbb{P}_D(\omega_i) \cdot \log(\mathbb{P}_K(\omega_i))
\end{equation}
Previously, we stated that a summary should aim, using only information from $D$, to offer the maximum amount of new information with respect to $K$.
$PI_K(D)$ can be understood as \emph{Potential Information} or maximum \emph{Informativeness}, the maximum amount of new information that a summary can extract from $D$ while knowing $K$.
% It can be viewed as the maximum \emph{Informativeness} available in $D$.
A summary $S$ cannot extract more than $PI_K(D)$ bits of information (if using only information from $D$).

\section{Experiments}
\label{sec:exp}
\subsection{Experimental setup}
To further illustrate the workings of the formula, we provide examples of experiments done with a simplistic choice for semantic units: words.
Even with simple assumptions $\theta_I$ is a meaningful quantity which correlates well with human judgments. \\

% We aim to investigate with different summarization tasks (Single Document, Multi-Document and Update summarization) if some parameters of the \(\alpha\)-Imp can model the human judgments. We are interested to know if and how these parameters differ from one task to another. We expect to observe 3 different notions of \(\alpha\)-Imp which cover the 3 different intuitions that humans have the 3 different summarization tasks. However, we should observe that this does not serve as a proper test for the theory because words are clearly poor approximations for the concept of semantic units. This is can reveal how far we can hope to go with this simplistic assumption. This serves as an illustration showing that the theory provides strong insights into applied summarization.

\noindent
\textbf{Data}: \\
We experiment with standard datasets for two different summarization tasks: generic and update multi-document summarization.

 We use two datasets from the Text Analysis Conference (TAC) shared task: TAC-2008 and TAC-2009.\footnote{\url{http://tac.nist.gov/2009/Summarization/}, \url{http://tac.nist.gov/2008/}}
% TAC-2008 and TAC-2009 contain 48 and 44 topics, respectively.
% Each topic consists of 10 news articles to be summarized in a maximum of 100 words.
% We use only the so-called initial documents (A documents) for the generic part.
In the update part, 10 new documents (B documents) are to be summarized assuming that the first 10 documents (A documents) have already been seen. The generic task consists in summarizing the initial document set (A).

For each topic, there are 4 human reference summaries and a manually created Pyramid set \cite{Nenkova:2007}. In both editions, all system summaries and the 4 reference summaries were manually evaluated by NIST assessors for readability, content selection (with Pyramid) and overall responsiveness.
At the time of the shared tasks, 57 systems were submitted to TAC-2008 and 55 to TAC-2009. \\

\noindent
\textbf{Setup and Assumptions}: \\
To keep the experiments simple and focused on illustrating the formulas, we make several simplistic assumptions. First, we choose words as semantic units and therefore texts are represented as frequency distributions over words. This assumption was already employed by previous works using information-theoretic tools for summarization \cite{Haghighi:2009}. While it is limiting, this remains a simple approximation letting us observe the quantities in action.

\(K, \alpha\) and $\beta$ are the parameters of the theory and their choice is subject to empirical investigation. Here, we make simple choices: for update summarization, \(K\) is the frequency distribution over words in the background documents (A). For generic summarization, \(K\) is the uniform probability distribution over all words from the source documents.
Furthermore, we use $\alpha = \beta = 1$.

% To keep the experiments simple and focused on illustrating the formulas, we make several simplistic assumptions. First, we choose words as semantic units and therefore texts are represented as frequency distributions over words. This assumption was already employed by previous works in summarization \cite{Haghighi:2009}. While it is limiting, this remains a simple approximation letting us observe the quantities in action.

% \(K, \alpha\) and $\beta$ are the parameters of the theory and their choice is subject to investigation. Here, we made simple choices: for update summarization, \(K\) is the frequency distribution over n-grams in the background documents (A). For generic summarization, \(K\) is the uniform probability distribution over all n-grams from the source documents.
% Furthermore, we use $\alpha = \beta = 1$.

\subsection{Correlation with humans}
First, we measure how well the different quantities correlate with human judgments. We compute the score of each system summary according to each quantity defined in the previous section: $Red$, $Rel$, $Inf$, $\theta_I(S, D, K)$. We then compute the correlations between these scores and the manual Pyramid scores. Indeed, each quantity is a summary scoring function and could, therefore, be evaluated based on its ability to correlate with human judgments \cite{Lin:2003:ncc}. Thus, we also report the performances of the summary scoring functions from several standard baselines:
\textbf{Edmundson} \cite{Edmundson1969} which scores sentences based on 4 methods: term frequency, presence of cue-words, overlap with title and position of the sentence.
\textbf{LexRank} \cite{Erkan2004} is a popular graph-based approach which scores sentences based on their centrality in a sentence similarity graph.
\textbf{ICSI} \cite{Gillick2009} extracts a summary by solving a maximum coverage problem considering the most frequent bigrams in the source documents. \textbf{KL} and \textbf{JS} \cite{Haghighi:2009} which measure the divergence between the distribution of words in the summary and in the sources.
Furthermore, we report two baselines from \newcite{louis:2014:P14-2} which account for background knowledge: KL\textsubscript{back} and JS\textsubscript{back} which measure the divergence between the distribution of the summary and the background knowledge $K$.
Further details concerning baseline scoring functions can be found in \cref{sec:baseline_details}.

        We measure the correlations with Kendall's $\tau$, a rank correlation metric which compares the orders induced by both scored lists.
        % Therefore, useful metrics should exhibit high NDCG scores.
        % \paragraph{Analysis}
        We report results for both generic and update summarization averaged over all topics for both datasets in \cref{tab:theta_corr}.
        \\
        % It is worth mentioning that with n-gram distributions as semantic units, our definition of \emph{Relevance} matches the objective function used in previous summarization systems \cite{Haghighi:2009,TUD-CS-20164649}.

        In general, the modelizations of \emph{Relevance} (based only on the sources) correlate better with human judgments than other quantities.
        Metrics accounting for background knowledge work better in the update scenario. This is not surprising as the background knowledge $K$ is more meaningful in this case (using the previous document set).

        We observe that JS divergence gives slightly better results than KL. Even though KL is more theoretically appealing, JS is smoother and usually works better in practice when distributions have different supports \cite{LouisN13}.

        Finally, $\theta_I$ significantly\footnote{at 0.01 with significance testing done with a t-test to compare two means} outperforms all baselines in both the generic and the update case. \emph{Red}, \emph{Rel} and \emph{Inf} are not particularly strong on their own, but combined together they yield a strong summary scoring function $\theta_I$. Indeed, each quantity models only one aspect of content selection, only together they form a strong signal for \emph{Importance}. \\

        % Moreover, the data might be biased because participating systems might have difficulties to address \emph{Redundancy} during the selection procedure.
        % This can explain why $Red$ performs well on these datasets. \emph{Redundancy} may be the main dimension of variation distinguishing the summaries submitted to the shared tasks.

        We need to be careful when interpreting these results because we made several strong assumptions: by choosing n-grams as semantic units and by choosing $K$ rather arbitrarily.
        Nevertheless, these are promising results. By investigating better text representations and more realistic $K$, we should expect even higher correlations.

        We provide a qualitative example on one topic in \cref{sec:example} with a visualization of $\mathbb{P}_{\frac{D}{K}}$ in comparison to reference summaries.\\
        % We already observe better correlation for $\theta_I$ in the update summarization scenario, which comes from a more natural choice of $K$.
        % Indeed, $K$ being the previous document set follows naturally from the task description.
        % In the generic case, the uninformative uniform distribution is a weaker approximation of background knowledge. \\

        \begin{table}[h]
                \small
                \centering
                % \resizebox{\columnwidth}{!}{
                \begin{tabular}{l|cc}
                \toprule
                                    & Generic   &   Update\\
				\midrule
                ICSI                &   .178      &  .139 \\
                Edm.                &   .215      &  .205 \\ %% 215
                LexRank             &   .201      &  .164 \\
                \midrule
                KL                  			 &   .204      &  .176 \\
                JS                  			 &   .225      &  .189 \\
                KL\textsubscript{back}           &   .110      &  .167 \\
                JS\textsubscript{back}           &   .066      &  .187 \\
                \midrule
                Red                 			&   .098      &  .096 \\
                Rel                 			&   .212      &  .192 \\
                Inf                 			&   .091      &  .086 \\
                \midrule
                $\theta_I$               		&   \textbf{.294} &  \textbf{.211} \\
                \bottomrule
                \end{tabular}
                \caption{Correlation of various information-theoretic quantities with human judgments measured by Kendall's $\tau$ on generic and update summarization.}\label{tab:theta_corr}
                % }
            \end{table}

\subsection{Comparison with Reference Summaries}
        Intuitively, the distribution $\mathbb{P}_{\frac{D}{K}}$ should be similar to the probability distribution $\mathbb{P}_R$ of the human-written reference summaries.
		% We now verify this hypothesis.
		% In this experiment, we verify that reference summaries have high $\theta_I$ scores.

        To verify this, we scored the system summaries and the reference summaries with $\theta_I$ and checked whether there is a significant difference between the two lists.\footnote{with standard $t$-test for comparing two related means.}
        % \footnote{We used the implementation from \url{https://www.scipy.org}}
        We found that $\theta_I$ scores reference summaries significantly higher than system summaries. The $p-$value, for the generic case, is $9.2\mathrm{e}{-6}$ and $1.1\mathrm{e}{-3}$ for the update case. Both are much smaller than the $1\mathrm{e}{-2}$ significance level.
		Therefore, $\theta_I$ is capable of distinguishing systems summaries from human written ones.
		For comparison, the best baseline (JS) has the following $p-$values: $8.2\mathrm{e}{-3}$ (Generic) and $4.5\mathrm{e}{-2}$ (Update). It does not pass the $1\mathrm{e}{-2}$ significance level for the update scenario.
        \\

\section{Conclusion and Future Work}
In this work, we argued for the development of theoretical models of \emph{Importance} and proposed one such framework.
Thus, we investigated a theoretical formulation of the notion of \emph{Importance}. In a framework rooted in information theory, we formalized several summary-related quantities like: \emph{Redundancy}, \emph{Relevance} and \emph{Informativeness}. \emph{Importance} arises as the notion unifying these concepts. More generally, \emph{Importance} is the measure that guides which choices to make when information must be discarded.
The introduced quantities generalize the intuitions that have previously been used in summarization research.
% Furthermore, under simplifying assumptions, we saw that the induced scoring function $\theta_I$ could correlate well with humans and results in a strong summarizer when combined with an appropriate optimizer.

Conceptually, it is straightforward to build a system out of $\theta_I$ once a semantic units representation and a $K$ have been chosen. A summarizer intends to extract or generate a summary maximizing $\theta_I$. This fits within the general optimization framework for summarization \cite{McDonald2007,peyrard-pyramid,Peyrard-objective}
% Indeed, $\theta_I$ is a summary scoring function which fits within the ($\theta$, $O$) framework.

The background knowledge and the choice of semantic units are free parameters of the theory. They are design choices which can be explored empirically by subsequent works. Our experiments already hint that strong summarizers can be developed from this framework.
Characters, character n-grams, morphemes, words, n-grams, phrases, and sentences do not actually qualify as semantic units. Even though previous works who relied on information theoretic motivation \cite{lin-EtAl:2006,Haghighi:2009,LouisN13,TUD-CS-20164649} used some of them as support for probability distributions, they are neither atomic nor independent. It is mainly because they are surface forms whereas semantic units are abstract and operate at the semantic level. However, they might serve as convenient approximations. Then, interesting research questions arise like \emph{Which granularity offers a good approximation of semantic units?} \emph{Can we automatically learn good approximations?} N-grams are known to be useful, but other granularities have rarely been considered together with information-theoretic tools.

For the background knowledge $K$, a promising direction would be to use the framework to actually learn it from data. In particular, one can apply supervised techniques to automatically search for $K$, $\alpha$ and $\beta$: finding the values of these parameters such that $\theta_I$ has the best correlation with human judgments.
By aggregating over many users and many topics one can find a generic $K$: what, on average, people consider as known when summarizing a document.
By aggregating over different people but in one domain, one can uncover a domain-specific $K$. Similarly, by aggregating over many topics for one person, one would find a personalized $K$.

These consistute promising research directions for future works.
% % This development is a humble starting point subject to be empirically tested and refined upon new empirical observations.

% %We hope this work encourages the community to develop more systems with automatic Pyramid in mind rather than solely focusing on developing systems improving ROUGE performances.

\section*{Acknowledgements}
This work was partly supported by the German Research Foundation (DFG) as part of the Research Training Group ``Adaptive Preparation of Information from Heterogeneous Sources'' (AIPHES) under grant No. GRK 1994/1, and via the German-Israeli Project Cooperation (DIP, grant No. GU 798/17-1).
We also thank the anonymous reviewers for their comments.

% \clearpage

\bibliographystyle{acl_natbib}
\bibliography{theo_summ}

\begin{thebibliography}{77}
\expandafter\ifx\csname natexlab\endcsname\relax\def\natexlab#1{#1}\fi

\bibitem[{Allahyari et~al.(2017)Allahyari, Pouriyeh, Assefi, Safaei, Trippe,
  Gutierrez, and Kochut}]{Allahyari:2017}
Mehdi Allahyari, Seyedamin Pouriyeh, Mehdi Assefi, Saeid Safaei, Elizabeth~D.
  Trippe, Juan~B. Gutierrez, and Krys Kochut. 2017.
\newblock \href {https://doi.org/10.14569/IJACSA.2017.081052} {{Text
  Summarization Techniques: A Brief Survey}}.
\newblock \emph{International Journal of Advanced Computer Science and
  Applications}, 8(10).

\bibitem[{Bao et~al.(2011)Bao, Basu, Dean, Partridge, Swami, Leland, and
  Hendler}]{bao2011towards}
Jie Bao, Prithwish Basu, Mike Dean, Craig Partridge, Ananthram Swami, Will
  Leland, and James~A Hendler. 2011.
\newblock {Towards a theory of semantic communication}.
\newblock In \emph{Network Science Workshop (NSW), 2011 IEEE}, pages 110--117.
  IEEE.

\bibitem[{Barzilay and Elhadad(1999)}]{barzilay1999using}
Regina Barzilay and Michael Elhadad. 1999.
\newblock {Using Lexical Chains for Text Summarization}.
\newblock \emph{{Advances in Automatic Text Summarization}}, pages 111--121.

\bibitem[{Blei et~al.(2003)Blei, Ng, and Jordan}]{blei2003latent}
David~M. Blei, Andrew~Y. Ng, and Michael~I. Jordan. 2003.
\newblock {Latent Dirichlet Allocation}.
\newblock \emph{{Journal of Machine Learning Research}}, 3:993--1022.

\bibitem[{Carbonell and Goldstein(1998)}]{Carbonell:1998}
Jaime Carbonell and Jade Goldstein. 1998.
\newblock \href {https://doi.org/10.1145/290941.291025} {{The Use of MMR,
  Diversity-based Reranking for Reordering Documents and Producing Summaries}}.
\newblock In \emph{Proceedings of the 21st Annual International ACM SIGIR
  Conference on Research and Development in Information Retrieval}, SIGIR '98,
  pages 335--336.

\bibitem[{Carnap and Bar-Hillel(1953)}]{Carnap}
Rudolf Carnap and Yehoshua Bar-Hillel. 1953.
\newblock \href {http://hdl.handle.net/1721.1/4821} {{An Outline of a Theory of
  Semantic Information}}.
\newblock \emph{British Journal for the Philosophy of Science.}, 4.

\bibitem[{Celikyilmaz and Hakkani-Tur(2010)}]{celikyilmaz:2010:ACL}
Asli Celikyilmaz and Dilek Hakkani-Tur. 2010.
\newblock \href {http://www.aclweb.org/anthology/P10-1084} {{A Hybrid
  Hierarchical Model for Multi-Document Summarization}}.
\newblock In \emph{{Proceedings of the 48th Annual Meeting of the Association
  for Computational Linguistics}}, pages 815--824, Uppsala, Sweden. Association
  for Computational Linguistics.

\bibitem[{Chali and Joty(2008)}]{chali2008improving}
Yllias Chali and Shafiq~R. Joty. 2008.
\newblock Improving the performance of the random walk model for answering
  complex questions.
\newblock In \emph{Proceedings of the 46th Annual Meeting of the Association
  for Computational Linguistics on Human Language Technologies: Short Papers},
  pages 9--12. Association for Computational Linguistics.

\bibitem[{Conroy et~al.(2006)Conroy, Schlesinger, and
  O'Leary}]{conroy-schlesinger-oleary:2006}
John~M. Conroy, Judith~D. Schlesinger, and Dianne~P. O'Leary. 2006.
\newblock \href {http://www.aclweb.org/anthology/P/P06/P06-2020}
  {{Topic-Focused Multi-Document Summarization Using an Approximate Oracle
  Score}}.
\newblock In \emph{Proceedings of the COLING/ACL 2006 Main Conference Poster
  Sessions}, pages 152--159, Sydney, Australia. Association for Computational
  Linguistics.

\bibitem[{Cruse(1986)}]{cruse86lexicalsemantics}
D.A. Cruse. 1986.
\newblock \emph{Lexical Semantics}.
\newblock Cambridge University Press, Cambridge, UK.

\bibitem[{Das and Martins(2010)}]{das2007survey}
Dipanjan Das and André F.~T. Martins. 2010.
\newblock {A Survey on Automatic Text Summarization}.
\newblock \emph{Literature Survey for the Language and Statistics II Course at
  CMU}.

\bibitem[{Daum{\'e} and Marcu(2002)}]{Daume:2002}
Hal Daum{\'e}, III and Daniel Marcu. 2002.
\newblock \href {https://doi.org/10.3115/1073083.1073159} {{A Noisy-channel
  Model for Document Compression}}.
\newblock In \emph{Proceedings of the 40th Annual Meeting on Association for
  Computational Linguistics}, pages 449--456.

\bibitem[{Daum{\'e}~III and Marcu(2006)}]{daume2006bayesian}
Hal Daum{\'e}~III and Daniel Marcu. 2006.
\newblock {Bayesian Query-Focused Summarization}.
\newblock In \emph{Proceedings of the 21st International Conference on
  Computational Linguistics and the 44th annual meeting of the Association for
  Computational Linguistics}, pages 305--312. Association for Computational
  Linguistics.

\bibitem[{Davis et~al.(2012)Davis, Conroy, and Schlesinger}]{davis2012occams}
Sashka~T. Davis, John~M. Conroy, and Judith~D. Schlesinger. 2012.
\newblock {OCCAMS--An Optimal Combinatorial Covering Algorithm for
  Multi-document Summarization}.
\newblock In \emph{{Proceeding of the 12th International Conference on Data
  Mining Workshops (ICDMW)}}, pages 454--463. IEEE.

\bibitem[{Deerwester et~al.(1990)Deerwester, Dumais, Furnas, Landauer, and
  Harshman}]{deerwester1990indexing}
Scott Deerwester, Susan~T. Dumais, George~W. Furnas, Thomas~K. Landauer, and
  Richard Harshman. 1990.
\newblock {Indexing by Latent Semantic Analysis}.
\newblock \emph{{Journal of the American Society for Information Science}},
  41(6):391--407.

\bibitem[{Delort and Alfonseca(2012)}]{Delort:2012}
Jean-Yves Delort and Enrique Alfonseca. 2012.
\newblock \href {http://dl.acm.org/citation.cfm?id=2380816.2380845} {{DualSum:
  A Topic-model Based Approach for Update Summarization}}.
\newblock In \emph{Proceedings of the 13th Conference of the European Chapter
  of the Association for Computational Linguistics}, pages 214--223.

\bibitem[{Dunning(1993)}]{dunning1993accurate}
Ted Dunning. 1993.
\newblock {Accurate Methods for the Statistics of Surprise and Coincidence}.
\newblock \emph{Computational linguistics}, 19(1):61--74.

\bibitem[{Edmundson(1969)}]{Edmundson1969}
H.~P. Edmundson. 1969.
\newblock \href {https://doi.org/10.1145/321510.321519} {{New Methods in
  Automatic Extracting}}.
\newblock \emph{{Journal of the Association for Computing Machinery}},
  16(2):264--285.

\bibitem[{Erk(2010)}]{erk2010word}
Katrin Erk. 2010.
\newblock {What is Word Meaning, Really? (and How Can Distributional Models
  Help Us Describe It?)}.
\newblock In \emph{{Proceedings of the 2010 workshop on geometrical models of
  natural language semantics}}, pages 17--26. Association for Computational
  Linguistics.

\bibitem[{Erkan and Radev(2004)}]{Erkan2004}
G\"{u}nes Erkan and Dragomir~R. Radev. 2004.
\newblock {LexRank: Graph-based Lexical Centrality As Salience in Text
  Summarization}.
\newblock \emph{{Journal of Artificial Intelligence Research}}, pages 457--479.

\bibitem[{Fillmore(1976)}]{Fillmore}
Charles~J. Fillmore. 1976.
\newblock \href {https://doi.org/10.1111/j.1749-6632.1976.tb25467.x} {{Frame
  Semantics And the Nature of Language}}.
\newblock \emph{Annals of the New York Academy of Sciences}, 280(1):20--32.

\bibitem[{Floridi(2009)}]{floridi2009philosophical}
Luciano Floridi. 2009.
\newblock {Philosophical Conceptions of Information}.
\newblock In \emph{Formal Theories of Information}, pages 13--53. Springer.

\bibitem[{Fujishige(2005)}]{opac-b1135069}
Satoru Fujishige. 2005.
\newblock \emph{{Submodular functions and optimization}}.
\newblock Annals of discrete mathematics. Elsevier, Amsterdam, Boston, Paris.

\bibitem[{G\'{a}bor et~al.(2017)G\'{a}bor, Zargayouna, Tellier, Buscaldi, and
  Charnois}]{gabor-EtAl}
Kata G\'{a}bor, Haifa Zargayouna, Isabelle Tellier, Davide Buscaldi, and
  Thierry Charnois. 2017.
\newblock \href {https://www.aclweb.org/anthology/D17-1193} {{Exploring Vector
  Spaces for Semantic Relations}}.
\newblock In \emph{Proceedings of the 2017 Conference on Empirical Methods in
  Natural Language Processing}, pages 1814--1823, Copenhagen, Denmark.
  Association for Computational Linguistics.

\bibitem[{Gillick and Favre(2009)}]{Gillick2009}
Dan Gillick and Benoit Favre. 2009.
\newblock {A Scalable Global Model for Summarization}.
\newblock In \emph{Proceedings of the Workshop on Integer Linear Programming
  for Natural Language Processing}, pages 10--18, Boulder, Colorado.
  Association for Computational Linguistics.

\bibitem[{Hachey et~al.(2006)Hachey, Murray, and
  Reitter}]{hachey2006dimensionality}
Ben Hachey, Gabriel Murray, and David Reitter. 2006.
\newblock {Dimensionality Reduction Aids Term Co-Occurrence Based
  Multi-Document Summarization}.
\newblock In \emph{{Proceedings of the Workshop on Task-Focused Summarization
  and Question Answering}}, pages 1--7. Association for Computational
  Linguistics.

\bibitem[{Haghighi and Vanderwende(2009)}]{Haghighi:2009}
Aria Haghighi and Lucy Vanderwende. 2009.
\newblock \href {http://www.aclweb.org/anthology/N/N09/N09-1041} {{Exploring
  Content Models for Multi-document Summarization}}.
\newblock In \emph{{Proceedings of Human Language Technologies: The 2009 Annual
  Conference of the North American Chapter of the Association for Computational
  Linguistics}}, pages 362--370.

\bibitem[{Harabagiu and Lacatusu(2005)}]{Harabagiu:2005}
Sanda Harabagiu and Finley Lacatusu. 2005.
\newblock \href {https://doi.org/10.1145/1076034.1076071} {{Topic Themes for
  Multi-document Summarization}}.
\newblock In \emph{{Proceedings of the 28th Annual International ACM SIGIR
  Conference on Research and Development in Information Retrieval}}, pages
  202--209.

\bibitem[{Harris(1954)}]{harris54}
Zellig Harris. 1954.
\newblock Distributional structure.
\newblock \emph{Word}, 10:146--162.

\bibitem[{He et~al.(2012)He, Chen, Bu, Wang, Zhang, Cai, and
  He}]{he2012document}
Zhanying He, Chun Chen, Jiajun Bu, Can Wang, Lijun Zhang, Deng Cai, and Xiaofei
  He. 2012.
\newblock {Document Summarization Based on Data Reconstruction}.
\newblock In \emph{{Proceeding of the Twenty-Sixth Conference on Artificial
  Intelligence}}.

\bibitem[{Hong et~al.(2014)Hong, Conroy, Favre, Kulesza, Lin, and
  Nenkova}]{HONG14}
Kai Hong, John Conroy, benoit Favre, Alex Kulesza, Hui Lin, and Ani Nenkova.
  2014.
\newblock {A Repository of State of the Art and Competitive Baseline Summaries
  for Generic News Summarization}.
\newblock In \emph{{Proceedings of the Ninth International Conference on
  Language Resources and Evaluation (LREC'14)}}, pages 1608--1616, Reykjavik,
  Iceland.

\bibitem[{Jaynes(1957)}]{PhysRev.106.620}
Edwin~T. Jaynes. 1957.
\newblock \href {https://doi.org/10.1103/PhysRev.106.620} {{Information Theory
  and Statistical Mechanics}}.
\newblock \emph{Physical Review}, 106:620--630.

\bibitem[{Kedzie et~al.(2018)Kedzie, McKeown, and Daume~III}]{Kedzie}
Chris Kedzie, Kathleen McKeown, and Hal Daume~III. 2018.
\newblock \href {http://aclweb.org/anthology/D18-1208} {{Content Selection in
  Deep Learning Models of Summarization}}.
\newblock In \emph{Proceedings of the 2018 Conference on Empirical Methods in
  Natural Language Processing}, pages 1818--1828. Association for Computational
  Linguistics.

\bibitem[{Kullback and Leibler(1951)}]{kullback1951}
Solomon Kullback and Richard~A. Leibler. 1951.
\newblock \href {https://doi.org/10.1214/aoms/1177729694} {{On Information and
  Sufficiency}}.
\newblock \emph{The Annals of Mathematical Statistics}, 22(1):79--86.

\bibitem[{Lavrenko(2008)}]{lavrenko2008generative}
Victor Lavrenko. 2008.
\newblock \emph{A generative theory of relevance}, volume~26.
\newblock Springer Science \& Business Media.

\bibitem[{Leskovec et~al.(2005)Leskovec, Milic-Frayling, and
  Grobelnik}]{leskovec2005impact}
Jure Leskovec, Natasa Milic-Frayling, and Marko Grobelnik. 2005.
\newblock {Impact of Linguistic Analysis on the Semantic Graph Coverage and
  Learning of Document Extracts}.
\newblock In \emph{{Proceedings of the National Conference on Artificial
  Intelligence}}, pages 1069--1074.

\bibitem[{Li et~al.(2015)Li, Bing, Lam, Li, and Liao}]{Li:2015:RMS}
Piji Li, Lidong Bing, Wai Lam, Hang Li, and Yi~Liao. 2015.
\newblock \href {http://dl.acm.org/citation.cfm?id=2832415.2832426}
  {{Reader-Aware Multi-document Summarization via Sparse Coding}}.
\newblock In \emph{{Proceedings of the 24th International Conference on
  Artificial Intelligence }}, pages 1270--1276.

\bibitem[{Lin et~al.(2006)Lin, Cao, Gao, and Nie}]{lin-EtAl:2006}
Chin-Yew Lin, Guihong Cao, Jianfeng Gao, and Jian-Yun Nie. 2006.
\newblock \href {http://www.aclweb.org/anthology/N/N06/N06-1059} {{An
  Information-Theoretic Approach to Automatic Evaluation of Summaries}}.
\newblock In \emph{Proceedings of the Human Language Technology Conference at
  NAACL}, pages 463--470, New York City, USA.

\bibitem[{Lin and Hovy(2003)}]{Lin:2003:ncc}
Chin-Yew Lin and Eduard Hovy. 2003.
\newblock \href {https://doi.org/10.3115/1073445.1073465} {{Automatic
  Evaluation of Summaries Using N-gram Co-occurrence Statistics}}.
\newblock In \emph{{Proceedings of the 2003 Conference of the North American
  Chapter of the Association for Computational Linguistics on Human Language
  Technology}}, volume~1, pages 71--78.

\bibitem[{Lin and Bilmes(2011)}]{LinB11}
Hui Lin and Jeff~A. Bilmes. 2011.
\newblock {A Class of Submodular Functions for Document Summarization}.
\newblock In \emph{{Proceedings of the 49th Annual Meeting of the Association
  for Computational Linguistics (ACL)}}, pages 510--520, Portland, Oregon.

\bibitem[{Liu et~al.(2015)Liu, Yu, and Deng}]{Liu:2015:MSB}
He~Liu, Hongliang Yu, and Zhi-Hong Deng. 2015.
\newblock \href {http://dl.acm.org/citation.cfm?id=2887007.2887035}
  {{Multi-document Summarization Based on Two-level Sparse Representation
  Model}}.
\newblock In \emph{{Proceedings of the Twenty-Ninth AAAI Conference on
  Artificial Intelligence}}, pages 196--202.

\bibitem[{Louis(2014)}]{louis:2014:P14-2}
Annie Louis. 2014.
\newblock \href {http://www.aclweb.org/anthology/P14-2055} {{A Bayesian Method
  to Incorporate Background Knowledge during Automatic Text Summarization}}.
\newblock In \emph{Proceedings of the 52nd Annual Meeting of the Association
  for Computational Linguistics (Volume 2: Short Papers)}, pages 333--338,
  Baltimore, Maryland.

\bibitem[{Louis and Nenkova(2013)}]{LouisN13}
Annie Louis and Ani Nenkova. 2013.
\newblock \href {https://doi.org/10.1162/COLI\_a\_00123} {{Automatically
  Assessing Machine Summary Content Without a Gold Standard}}.
\newblock \emph{Computational Linguistics}, 39(2):267--300.

\bibitem[{Luhn(1958)}]{Luhn}
Hans~Peter Luhn. 1958.
\newblock {The Automatic Creation of Literature Abstracts}.
\newblock \emph{{IBM Journal of Research Development}}, 2:159--165.

\bibitem[{Ma et~al.(2016)Ma, Deng, and Yang}]{ma:2016:COLING}
Shulei Ma, Zhi-Hong Deng, and Yunlun Yang. 2016.
\newblock \href {http://aclweb.org/anthology/C16-1143} {{An Unsupervised
  Multi-Document Summarization Framework Based on Neural Document Model}}.
\newblock In \emph{{Proceedings of COLING 2016, the 26th International
  Conference on Computational Linguistics: Technical Papers}}, pages
  1514--1523. The COLING 2016 Organizing Committee.

\bibitem[{Mani(1999)}]{Maybury:1999}
Inderjeet Mani. 1999.
\newblock \emph{Advances in Automatic Text Summarization}.
\newblock MIT Press, Cambridge, MA, USA.

\bibitem[{Mani and Bloedorn(1997)}]{Mani1997}
Inderjeet Mani and Eric Bloedorn. 1997.
\newblock {Multi-document Summarization by Graph Search and Matching}.
\newblock In \emph{{Proceedings of the Fourteenth National Conference on
  Artificial Intelligence and Ninth Conference on Innovative Applications of
  Artificial Intelligence}}, pages 622--628, Providence, Rhode Island. AAAI
  Press.

\bibitem[{McDonald(2007)}]{McDonald2007}
Ryan McDonald. 2007.
\newblock \href {https://doi.org/10.1007/978-3-540-71496-5\_51} {{A Study of
  Global Inference Algorithms in Multi-document Summarization}}.
\newblock In \emph{{Proceedings of the 29th European Conference on Information
  Retrieval Research}}, pages 557--564.

\bibitem[{McKeown et~al.(1999)McKeown, Klavans, Hatzivassiloglou, Barzilay, and
  Eskin}]{McKeown:1999:TMS}
Kathleen~R. McKeown, Judith~L. Klavans, Vasileios Hatzivassiloglou, Regina
  Barzilay, and Eleazar Eskin. 1999.
\newblock \href {http://dl.acm.org/citation.cfm?id=315149.315355} {{Towards
  Multidocument Summarization by Reformulation: Progress and Prospects}}.
\newblock In \emph{{Proceedings of the Sixteenth National Conference on
  Artificial Intelligence and the Eleventh Innovative Applications of
  Artificial Intelligence Conference Innovative Applications of Artificial
  Intelligence}}, pages 453--460.

\bibitem[{Mihalcea and Tarau(2004)}]{mihalcea2004textrank}
Rada Mihalcea and Paul Tarau. 2004.
\newblock Textrank: Bringing order into text.
\newblock In \emph{Proceedings of the 2004 conference on empirical methods in
  natural language processing}.

\bibitem[{Mikolov et~al.(2013{\natexlab{a}})Mikolov, Chen, Corrado, and
  Dean}]{abs-1301-3781}
Tomas Mikolov, Kai Chen, Greg Corrado, and Jeffrey Dean. 2013{\natexlab{a}}.
\newblock \href {http://dblp.uni-trier.de/db/journals/corr/corr1301.html}
  {{Efficient Estimation of Word Representations in Vector Space}}.
\newblock \emph{CoRR}, abs/1301.3781.

\bibitem[{Mikolov et~al.(2013{\natexlab{b}})Mikolov, Sutskever, Chen, Corrado,
  and Dean}]{word2vec}
Tomas Mikolov, Ilya Sutskever, Kai Chen, Greg~S. Corrado, and Jeff Dean.
  2013{\natexlab{b}}.
\newblock \href {https://arxiv.org/pdf/1310.4546.pdf} {{Distributed
  representations of words and phrases and their compositionality}}.
\newblock In \emph{{Advances in Neural Information Processing Systems}}, pages
  3111--3119, Lake Tahoe, Nevada, USA.

\bibitem[{Montague(1970)}]{Montague1970-MONEAA-2}
Richard Montague. 1970.
\newblock English as a formal language.
\newblock In Bruno Visentini, editor, \emph{Linguaggi nella societa e nella
  tecnica}, pages 188--221. Edizioni di Communita.

\bibitem[{Nenkova and McKeown(2012)}]{Nenkova2012}
Ani Nenkova and Kathleen McKeown. 2012.
\newblock {A Survey of Text Summarization Techniques}.
\newblock \emph{Mining Text Data}, pages 43--76.

\bibitem[{Nenkova et~al.(2007)Nenkova, Passonneau, and McKeown}]{Nenkova:2007}
Ani Nenkova, Rebecca Passonneau, and Kathleen McKeown. 2007.
\newblock {The Pyramid Method: Incorporating Human Content Selection Variation
  in Summarization Evaluation}.
\newblock \emph{ACM Transactions on Speech and Language Processing (TSLP)},
  4(2).

\bibitem[{Nenkova et~al.(2006)Nenkova, Vanderwende, and
  McKeown}]{Nenkova:2006:CCS}
Ani Nenkova, Lucy Vanderwende, and Kathleen McKeown. 2006.
\newblock \href {https://doi.org/10.1145/1148170.1148269} {{A Compositional
  Context Sensitive Multi-document Summarizer: Exploring the Factors That
  Influence Summarization}}.
\newblock In \emph{{Proceedings of the 29th Annual International ACM SIGIR
  Conference on Research and Development in Information Retrieval}}, SIGIR '06,
  pages 573--580.

\bibitem[{Peyrard and Eckle-Kohler(2016)}]{TUD-CS-20164649}
Maxime Peyrard and Judith Eckle-Kohler. 2016.
\newblock \href {http://aclweb.org/anthology/C16-1024} {{A General Optimization
  Framework for Multi-Document Summarization Using Genetic Algorithms and Swarm
  Intelligence}}.
\newblock In \emph{{Proceedings of the 26th International Conference on
  Computational Linguistics (COLING)}}, pages 247 -- 257.

\bibitem[{Peyrard and Eckle-Kohler(2017{\natexlab{a}})}]{peyrard-principled}
Maxime Peyrard and Judith Eckle-Kohler. 2017{\natexlab{a}}.
\newblock \href {http://tubiblio.ulb.tu-darmstadt.de/104586/} {A principled
  framework for evaluating summarizers: Comparing models of summary quality
  against human judgments}.
\newblock In \emph{Proceedings of the 55th Annual Meeting of the Association
  for Computational Linguistics (ACL 2017)}, volume Volume 2: Short Papers,
  pages 26--31. Association for Computational Linguistics.

\bibitem[{Peyrard and Eckle-Kohler(2017{\natexlab{b}})}]{peyrard-pyramid}
Maxime Peyrard and Judith Eckle-Kohler. 2017{\natexlab{b}}.
\newblock \href {http://tubiblio.ulb.tu-darmstadt.de/104585/} {Supervised
  learning of automatic pyramid for optimization-based multi-document
  summarization}.
\newblock In \emph{Proceedings of the 55th Annual Meeting of the Association
  for Computational Linguistics (ACL 2017)}, volume Volume 1: Long Papers,
  pages 1084--1094. Association for Computational Linguistics.

\bibitem[{Peyrard and Gurevych(2018)}]{Peyrard-objective}
Maxime Peyrard and Iryna Gurevych. 2018.
\newblock \href {http://tubiblio.ulb.tu-darmstadt.de/97908/} {Objective
  function learning to match human judgements for optimization-based
  summarization}.
\newblock In \emph{Proceedings of the 16th Annual Conference of the North
  American Chapter of the Association for Computational Linguistics: Human
  Language Technologies}, pages 654--660. Association for Computational
  Linguistics.

\bibitem[{Radev et~al.(2000)Radev, Jing, and Budzikowska}]{Radev2000}
Dragomir~R. Radev, Hongyan Jing, and Malgorzata Budzikowska. 2000.
\newblock {Centroid-based Summarization of Multiple Documents: Sentence
  Extraction, Utility-based Evaluation, and User Studies}.
\newblock In \emph{{Proceedings of the NAACL-ANLP Workshop on Automatic
  Summarization}}, volume~4, pages 21--30, Seattle, Washington.

\bibitem[{Shannon(1948)}]{Shannon48}
Claude~E. Shannon. 1948.
\newblock \href {https://doi.org/10.1002/j.1538-7305.1948.tb00917.x} {{A
  Mathematical Theory of Communication}}.
\newblock \emph{Bell Systems Technical Journal}, 27:623--656.

\bibitem[{Sipos et~al.(2012)Sipos, Shivaswamy, and Joachims}]{Sipos2012}
Ruben Sipos, Pannaga Shivaswamy, and Thorsten Joachims. 2012.
\newblock {Large-margin Learning of Submodular Summarization Models}.
\newblock In \emph{{Proceedings of the 13th Conference of the European Chapter
  of the Association for Computational Linguistics}}, pages 224--233, Avignon,
  France. Association for Computational Linguistics.

\bibitem[{Sparck~Jones(1972)}]{sparck1972statistical}
Karen Sparck~Jones. 1972.
\newblock {A Statistical Interpretation of Term Specificity and its Application
  in Retrieval}.
\newblock \emph{{Journal of documentation}}, 28(1):11--21.

\bibitem[{Tsvetkov(2014)}]{tsvetkov2014ke}
Victor~Yakovlevich Tsvetkov. 2014.
\newblock {The KE Shannon and L. Floridi's Amount of Information}.
\newblock \emph{{Life Science Journal}}, 11(11):667--671.

\bibitem[{Turian et~al.(2010)Turian, Ratinov, and Bengio}]{Turian:2010}
Joseph Turian, Lev Ratinov, and Yoshua Bengio. 2010.
\newblock \href {http://dl.acm.org/citation.cfm?id=1858681.1858721} {{Word
  Representations: A Simple and General Method for Semi-supervised Learning}}.
\newblock In \emph{Proceedings of the 48th Annual Meeting of the Association
  for Computational Linguistics}, pages 384--394.

\bibitem[{Turney and Pantel(2010)}]{turney2010frequency}
Peter~D Turney and Patrick Pantel. 2010.
\newblock {From Frequency to Meaning: Vector Space Models of Semantics}.
\newblock \emph{Journal of artificial intelligence research}, 37:141--188.

\bibitem[{Vanderwende et~al.(2007)Vanderwende, Suzuki, Brockett, and
  Nenkova}]{VANDERWENDE20071606}
Lucy Vanderwende, Hisami Suzuki, Chris Brockett, and Ani Nenkova. 2007.
\newblock \href {https://doi.org/https://doi.org/10.1016/j.ipm.2007.01.023}
  {{Beyond SumBasic: Task-focused Summarization with Sentence Simplification
  and Lexical Expansion}}.
\newblock \emph{{Information Processing \& Management}}, 43(6):1606--1618.

\bibitem[{Wan and Yang(2006)}]{wan2006improved}
Xiaojun Wan and Jianwu Yang. 2006.
\newblock {Improved Affinity Graph Based Multi-Document Summarization}.
\newblock In \emph{{Proceedings of the Human Language Technology Conference of
  the NAACL, Companion Volume: Short Papers}}, pages 181--184. Association for
  Computational Linguistics.

\bibitem[{Wang et~al.(2009)Wang, Zhu, Li, and Gong}]{wang2009multi}
Dingding Wang, Shenghuo Zhu, Tao Li, and Yihong Gong. 2009.
\newblock {Multi-document Summarization Using Sentence-based Topic Models}.
\newblock In \emph{{Proceedings of the ACL-IJCNLP 2009}}, pages 297--300.
  Association for Computational Linguistics.

\bibitem[{Weaver(1953)}]{weaver1953recent}
Warren Weaver. 1953.
\newblock {Recent Contributions to the Mathematical Theory of Communication}.
\newblock \emph{{ETC: A Review of General Semantics}}, pages 261--281.

\bibitem[{Wilson and Sperber(2008)}]{Relevance-theory}
Deirdre Wilson and Dan Sperber. 2008.
\newblock \href {https://doi.org/10.1002/9780470756959.ch27} {\emph{Relevance
  Theory}}, chapter~27. John Wiley and Sons, Ltd.

\bibitem[{Yao et~al.(2017)Yao, Wan, and Xiao}]{Yao2017}
Jin-ge Yao, Xiaojun Wan, and Jianguo Xiao. 2017.
\newblock \href {https://doi.org/10.1007/s10115-017-1042-4} {{Recent Advances
  in Document Summarization}}.
\newblock \emph{Knowledge and Information Systems}, 53(2):297--336.

\bibitem[{Yogatama et~al.(2015)Yogatama, Liu, and
  Smith}]{yogatama-liu-smith:2015:EMNLP}
Dani Yogatama, Fei Liu, and Noah~A. Smith. 2015.
\newblock \href {http://aclweb.org/anthology/D15-1228} {{Extractive
  Summarization by Maximizing Semantic Volume}}.
\newblock In \emph{Proceedings of the 2015 Conference on Empirical Methods in
  Natural Language Processing}, pages 1961--1966, Lisbon, Portugal.

\bibitem[{Zhang et~al.(2015)Zhang, Xia, Liu, and Wang}]{zhang:2015:NAACL}
Yang Zhang, Yunqing Xia, Yi~Liu, and Wenmin Wang. 2015.
\newblock \href {http://www.aclweb.org/anthology/N15-1136} {{Clustering
  Sentences with Density Peaks for Multi-document Summarization}}.
\newblock In \emph{{Proceedings of the 2015 Conference of the North American
  Chapter of the Association for Computational Linguistics: Human Language
  Technologies}}, pages 1262--1267, Denver, Colorado. Association for
  Computational Linguistics.

\bibitem[{Zhong(2017)}]{Zhong}
Yixin Zhong. 2017.
\newblock \href {https://doi.org/10.3390/IS4SI-2017-04000} {{A Theory of
  Semantic Information}}.
\newblock In \emph{Proceedings of the IS4SI 2017 Summit Digitalisation for a
  Sustainable Society}, 129.

\bibitem[{Zopf et~al.(2016)Zopf, Menc{\'i}a, and
  F{\"u}rnkranz}]{TUD-CS-2016-0154}
Markus Zopf, Eneldo~Loza Menc{\'i}a, and Johannes F{\"u}rnkranz. 2016.
\newblock \href {http://aclweb.org/anthology/K/K16/K16-1009.pdf} {{Beyond
  Centrality and Structural Features: Learning Information Importance for Text
  Summarization}}.
\newblock In \emph{Proceedings of the 20th SIGNLL Conference on Computational
  Natural Language Learning (CoNLL 2016)}, pages 84--94.

\end{thebibliography}

\clearpage

\appendix
\section{Details about Baseline Scoring Functions}
\label{sec:baseline_details}
    In the paper, we compare the summary scoring function $\theta_I$ against the summary scoring functions derived from several summarizers following the methodology from \newcite{peyrard-principled}. Here, we give explicit formulation of the baseline scoring functions.

    \noindent
    \textbf{Edmundson}: \cite{Edmundson1969} \\
        \newcite{Edmundson1969} presented a heuristic which scores sentences according to 4 different features:

        \begin{itemize}
            \item \textbf{Cue-phrases}: It is based on the hypothesis that the probable relevance of a sentence is affected by the presence of certain cue words such as 'significant' or 'important'. Bonus words have positive weights, stigma words have negative weights and all the others have no weight. The final score of the sentence is the sum of the weights of its words.
            \item \textbf{Key}: High-frequency content words are believed to be positively correlated with relevance \cite{Luhn}. Each word receives a weight based on its frequency in the document if it is not a stopword. The score of the sentence is also the sum of the weights of its words.
            \item \textbf{Title}: It measures the overlap between the sentence and the title.
            \item \textbf{Location}: It relies on the assumption that sentences appearing early or late in the source documents are more relevant.
        \end{itemize}

        By combining these scores with a linear combination, we can recognize the objective function:
        \begin{align}
            \theta_{Edm.}(S) &= \sum_{s \in S} \alpha_1 \cdot C(s) + \alpha_2 \cdot K(s) \\
            &+ \alpha_3 \cdot T(s) + \alpha_4 \cdot L(s)
        \end{align}
        The sum runs over sentences and $C, K, T$ and $L$ output the sentence scores for each method (Cue, Key, Title and Location). \\

    \noindent
    \textbf{ICSI}: \cite{Gillick2009} \\
        A global linear optimization that extracts a summary by solving a maximum coverage problem of the most frequent bigrams in the source documents. ICSI has been among the best systems in a classical ROUGE evaluation \cite{HONG14}.
        Here, the identification of the scoring function is trivial because it was originally formulated as an optimization task. If $c_i$ is the $i$-th bigram selected in the summary and $w_i$ is its weight computed from $D$, then:
        \begin{equation}
            \theta_{ICSI}(S) = \sum\limits_{c_i \in S} c_i \cdot w_i
        \end{equation}
        % The extraction strategy is an Integer Linear Program. Note that \newcite{LiQL13} have later refined this approach by learning the weights $w_i$ instead of using the document frequency of the bigram. \\

    \noindent
    \textbf{LexRank}: \cite{Erkan2004} \\
        This is a well-known graph-based approach. A similarity graph $G(V, E)$ is constructed where $V$ is the set of sentences and an edge $e_{ij}$ is drawn between sentences $v_i$ and $v_j$ if and only if the cosine similarity between them is above a given threshold. Sentences are scored according to their PageRank score in $G$. Thus, $\theta_{LexRank}$ is given by:
        \begin{equation}
            \theta_{LexRank}(S) = \sum\limits_{s \in S} PR_{G}(s)
        \end{equation}
        Here, $PR$ is the PageRank score of sentence $s$. \\

    \noindent
    \textbf{KL-Greedy}: \cite{Haghighi:2009} \\
        In this approach, the summary should minimize the Kullback-Leibler (KL) divergence between the word distribution of the summary $S$ and the word distribution of the documents $D$ (i.e., $\theta_{KL} = -KL$):
        \begin{align}
            \theta_{KL}(S) = -KL(S||D) \\
            = -\sum\limits_{g \in S} \mathbb{P}_S(g) \log  \frac{\mathbb{P}_S(g)}{\mathbb{P}_D(g)}
        \end{align}

        $\mathbb{P}_X(w)$ represents the frequency of the word (or n-gram) $w$ in the text $X$. The minus sign indicates that KL should be lower for better summaries. Indeed, we expect a good system summary to exhibit a similar probability distribution of n-grams as the sources.

        Alternatively, the Jensen-Shannon (JS) divergence can be used instead of KL.
        Let $M$ be the average word frequency distribution of the candidate summary $S$ and the source documents $D$ distribution:
        \begin{equation}
            \forall g \in S, \ \mathbb{P}_M(g) = \frac{1}{2}(\mathbb{P}_S(g) + \mathbb{P}_{D}(g))
        \end{equation}

        Then, the formula for JS is given by:
        \begin{align}
            \theta_{JS}(S) &= -JS(S||D) \\
            &= \frac{1}{2}\left(KL(S||M) + KL(D||M)\right)
        \end{align}
        % The negative sign indicates that lower divergences are better than higher ones.
        Within our framework, the KL divergence acts as the unification of \emph{Relevance} and \emph{Redundancy} when semantic units are bigrams.\\

\section{Proof of Theorem 1}
\label{sec:proof}
Let $\Omega$ be the set of semantic units. The notation $\omega_i$ represents one unit.
Let $\mathbb{P}_T$, and $\mathbb{P}_K$ be the text representations of the source documents and background knowledge as probability distributions over semantic units.

We note $t_i = \mathbb{P}_T(\omega_i)$, the probability of the unit $\omega_i$ in the source $T$. Similarly, we note $k_i = \mathbb{P}_K(\omega_i)$. We seek a function $f$ unifying $T$ and $K$ such that: $f(\omega_i) = f(t_i, k_i)$.

We remind the simple requirements that $f$ should satisfy:
\begin{itemize}
\item Informativeness: $\forall i \neq j$, if $t_i = t_j$ and $k_i > k_j$ then $f(t_i, k_i) < f(t_j, k_j)$
\item Relevance: $\forall i \neq j$, if $t_i > t_j$ and $k_i = k_j$ then $f(t_i, k_i) > f(t_j, k_j)$
\item Additivity: $I(f(t_i, k_i)) \equiv \alpha I(t_i) + \beta I(k_i)$ ($I$ is the information measure from Shannon's theory \cite{Shannon48})
\item Normalization: $\sum\limits_{i} f(t_i, k_i) = 1$
\end{itemize}

Theorem \ref{th:main_th} states that the functions satisfying the previous requirements are:
\begin{equation}
\begin{split}
\mathbb{P}_{\frac{T}{K}}(\omega_i) = \frac{1}{C} \cdot \frac{t_i^{\alpha}}{k_i^{\beta}} \\
C = \sum\limits_{i} \frac{t_i^{\alpha}}{k_i^{\beta}} \text{ , }\alpha, \beta \in \mathbb{R}^{+}
\end{split}
\end{equation}
with $C$ the normalizing constant.

\begin{proof}
The information function defined by \newcite{Shannon48} is the logarithm: $I = \log$.
Then, the \emph{Additivity} criterion can be written:
\begin{equation}
\log(f(t_i, k_i)) = \alpha \log(t_i) + \beta \log(k_i) + A
\end{equation}
with $A$ a constant independent of $t_i$ and $k_i$

Since $\log$ is monotonous and increasing, the \emph{Informativeness} and \emph{Additivity} criteria can be combined:

$\forall i \neq j$, if $t_i = t_j$ and $k_i > k_j$ then:
\begin{align*}
\log f(t_i, k_i) & < \log f(t_j, k_j) \\
\alpha \log(t_i) + \beta \log(k_i) & < \alpha \log(t_j) + \beta \log(k_j)  \\
\beta \log(k_i) &< \beta \log(k_j)  \\
\text{But } k_i > k_j \text{, therefore: } \\
\beta & < 0
\end{align*}
For clarity, we can now use $-\beta$ with $\beta \in \mathbb{R}^{+}$.

Similarly, we can combine the \emph{Relevance} and \emph{Additivity} criteria:
$\forall i \neq j$, if $t_i > t_j$ and $k_i = k_j$ then:

\begin{align*}
\log f(t_i, k_i) & > \log f(t_j, k_j) \\
\alpha \log(t_i) + \beta \log(k_i) & > \alpha \log(t_j) + \beta \log(k_j) \\
\alpha \log(t_i) & > \alpha \log(t_j) \\
\text{But } t_i > t_j \text{, therefore: } \\
\alpha & > 0
\end{align*}

Then, we have the following form from the \emph{Additivity} criterion:
\begin{align*}
\log f(t_i, k_i) & = \alpha \log(t_i) - \beta \log(k_i) + A \\
f(t_i, k_i) & = e^{A} e^{[\alpha \log(t_i) - \beta \log(k_i)]}  \\
f(t_i, k_i) & = e^{A} \frac{t_i^{\alpha}}{k_i^{\beta}} x \\
\end{align*}

Finally, the \emph{Normalization} constraint specifies the constant $e^{A}$:
\begin{align*}
C & = \frac{1}{e^A} \\
\text{and } C & = \sum\limits_{i} \frac{t_i^{\alpha}}{k_i^{\beta}} \\
\text{then: } A & = - \log (\sum\limits_{i} \frac{t_i^{\alpha}}{k_i^{\beta}})
\end{align*}
\end{proof}

\section{Example}
\label{sec:example}

        As an example, for one selected topic of TAC-2008 update track, we computed the $\mathbb{P}_{\frac{D}{K}}$ and compare it to the distribution of the $4$ reference summaries.

        \begin{figure*}
        \centering
        \includegraphics[width=0.9\linewidth]{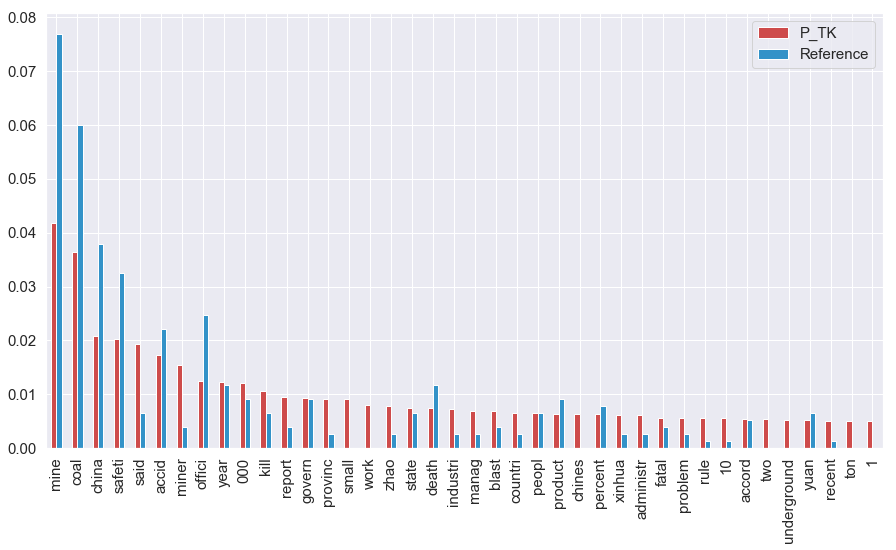}
          \caption{Example of $\mathbb{P}_{\frac{D}{K}}$ in comparison to the word distribution of reference summaries for one topic of TAC-2008 (D0803).}
        \label{fig:example_p_tk}
        \end{figure*}

        We report the two distributions together in \cref{fig:example_p_tk}. For visibility, only the top $50$ words according to $\mathbb{P}_{\frac{D}{K}}$ are considered. However, we observe a good match between the distribution of the reference summaries and the \emph{ideal} distribution as defined by $\mathbb{P}_{\frac{D}{K}}$.

        Furthermore, the most desired words according to $\mathbb{P}_{\frac{D}{K}}$ make sense. This can be seen by looking at one of the human-written reference summary of this topic:

        \begin{quote}
        \textbf{Reference summary for topic D0803} \\
        \emph{China sacrificed coal mine safety in its massive demand for energy. Gas explosions, flooding, fires, and cave-ins cause most accidents. The mining industry is riddled with corruption from mining officials to owners. Officials are often illegally invested in mines and ignore safety procedures for production. South Africa recently provided China with information on mining safety and technology during a conference. China is beginning enforcement of safety regulations. Over 12,000 mines have been ordered to suspend operations and 4,000 others ordered closed. This year 4,228 miners were killed in 2,337 coal mine accidents. China's mines are the most dangerous worldwide.}
        \end{quote}

% \subsection{Connection between $I_K$ and the Multinomial Generative Story}: \\

% Suppose $T$ the source is a semantic source emitting semantic units with probabilities: $t_1, \dots t_m$.
% And we observe the summary $S$ with actual probabilities: $\frac{s_1}{n}, \dots, \frac{s_m}{n}$, where $n = \sum s_i$ and $\min s_i = 1$.

% Assume $S$ is being drawn from $T$ according to a multinomial distribution:
% \begin{equation}
% P(X_1 = s_1, \dots, X_m = s_m) = \frac{\Gamma(\sum s_i +1)}{\prod \Gamma(s_i + 1)} \prod p_i^{s_i}
% \end{equation}

% This probability corresponds to the probability of observing the $w_i$ units emitted for $s_i$ times in the summary if the probability of emitting is $p_i$. When we observe a summary $S$ drawn according to the multinomial from the source $T$, we should expect to see a high log-likelihood:

% \begin{align*}
% l(T|S) & = \log \Gamma(\sum s_i +1) - \sum \log \Gamma(s_i + 1) + \sum s_i \log p_i \\
%        & = C - \sum s_i \log s_i - s_i + -CE(S,T) \\
%        & = C + H(S) - \sum s_i + -CE(S,T) \\
%         & \equiv -Red(S) + Rel(S,T) \\
% \end{align*}

% We can do the approximation of the gamma function because $s_i \geq 1$ and the stirling approximation is valid.

% If we use the prior with a multinomial $\frac{1}{k_i}$, then we observe that the full formula is maximizing the MAP
% \begin{equation}
% l_{MAP}(T|S) \equiv -Red(S) + Rel(S,T) + Inf(S,K)
% \end{equation}

% Furthermore, this formula is also the MLE when the multinomial is done with the importance encoding distribution. \\

% \section{Experiments}
% \label{sec:experiments}
% \input{experiments}

\end{document}